\documentclass[10pt,twocolumn,letterpaper]{article}

\usepackage[pagenumbers]{wacv} 
\usepackage[accsupp]{axessibility}
\usepackage{graphicx}
\usepackage{amsmath}
\usepackage{amssymb}
\usepackage{booktabs}
\usepackage{multirow}
\usepackage{color}

\usepackage[pagebackref,breaklinks,colorlinks]{hyperref}

\usepackage[capitalize]{cleveref}
\crefname{section}{Sec.}{Secs.}
\Crefname{section}{Section}{Sections}
\Crefname{table}{Table}{Tables}
\crefname{table}{Tab.}{Tabs.}

\def\wacvPaperID{2117} 
\def\confName{WACV}
\def\confYear{2025}

\begin{document}

\title{HeightMapNet: Explicit Height Modeling for End-to-End HD Map Learning}

\author{Wenzhao Qiu,~~ Shanmin Pang\thanks{Corresponding author}, ~~Hao Zhang, ~~Jianwu Fang, ~~Jianru Xue\\
Xi'an Jiaotong University\\
{\tt\small \{a2801551754@stu., pangsm@, belief@stu., fangjianwu@, jrxue@\}xjtu.edu.cn}
\thanks{This work was supported  by the National Key R\&D Program of China under Grant 2022ZD0117903.}
}
\maketitle

\begin{abstract}
    Recent advances in high-definition (HD) map construction from surround-view images have highlighted their cost-effectiveness in deployment. However, prevailing techniques often fall short in accurately extracting and utilizing road features, as well as in the implementation of view transformation. In response, we introduce HeightMapNet, a novel framework that establishes a dynamic relationship between image features and road surface height distributions. By integrating height priors, our approach refines the accuracy of Bird’s-Eye-View (BEV) features beyond conventional methods. HeightMapNet also  introduces a foreground-background separation network that sharply distinguishes between critical road elements and extraneous background components, enabling precise focus on detailed road micro-features. Additionally, our method leverages multi-scale features within the BEV space, optimally utilizing spatial geometric information to boost model performance. HeightMapNet has shown exceptional results on the challenging nuScenes and Argoverse 2 datasets, outperforming several widely recognized approaches. The code will be available  at \url{https://github.com/adasfag/HeightMapNet/}.
\end{abstract}

\section{Introduction}
\label{sec:intro}

In the domain of autonomous driving, the ability to accurately and comprehensively interpret the environmental context surrounding the ego vehicle is paramount for ensuring safe and effective operational decisions. Surround-view methods, recognized for their cost-efficiency and broad applicability, have garnered significant advancements in this area.
Current  techniques predominantly fall into two categories based on their intermediary representations: sparse query-based methods and 
 dense BEV-based methods. 
Drawing inspiration from the DETR architecture~\cite{carion2020end,detr3d}, sparse query-based approaches~\cite{liu2023petrv2,2305.14018,2211.10581} utilize learnable global queries representing detection elements, refined through interactions with surround-view image features. Although this strategy effectively controls the proliferation of queries, its reliance on static global queries limits adaptability in dynamic environments, often resulting in detection inaccuracies at extended distances. Conversely, BEV-based methods convert Perspective View (PV) into BEV representations using a perspective transformation module.
Subsequent feature processing is performed via a map detector head.  BEV methods~\cite{yu2023scalablemap, li2022hdmapnet, MapTR, maptrv2, liu2022vectormapnet} demonstrate state-of-the-art performance in online mapping, and  have recently  dominated the field.

Existing BEV methods typically utilize enhanced LSS \cite{philion2020lift} or attention mechanism~\cite{vaswani2017attention} as baselines for the perspective transformation module. LSS-based methods \cite{maptrv2,li2024dtclmapperdualtemporalconsistent} often require auxiliary losses to accelerate detector convergence speed, while attention-based methods \cite{yu2023scalablemap,ding2023pivotnet,Qiao_2023_CVPR,qiao2023machmap,liu2024leveraging,chen2024maptracker} usually need additional modules to improve the output BEV features. These methods normally overlook the vertical dimension of road features during the PV-to-BEV transformation, thus compromising their capacity to delineate complex environmental details accurately. 
Additionally, most existing studies~\cite{li2022bevformer, MapTR, liu2023petrv2} inadequately address the challenge of  filtering non-critical elements such as the sky and other extraneous background features during the processing of image features from multi-view inputs. This oversight in selective background filtration foregoes crucial noise reduction opportunities, consequently rendering the models vulnerable to disruptions caused by irrelevant data. Such disruptions significantly compromise the accuracy and reliability of the resultant perceptual outputs. Moreover, while prevailing researches~\cite{li2022bevformer, MapTR, maptrv2} tend to concentrate on exploiting single-layer image features for computational effectiveness, it largely neglects the  benefits and possibilities afforded by multi-scale feature fusion within the BEV space. This limitation undermines the model’s effectiveness in navigating complex road environments.

To address these challenges, we propose a new view transformation paradigm that forges a nuanced relationship between image features and road surface height distributions, integrating height priors to refine the accuracy of BEV features. By explicitly modeling ground height during vehicle operation, the capacity to delineate complex environmental details is significantly improved. In addition, we develop a foreground-background separation network with self-supervised learning. This network optimizes the extraction and utilization of road features, effectively  minimizing non-critical background elements, thereby enhancing the clarity and quality of the input features. Such enhancements substantially improve the reliability of the perceptual outcomes. Furthermore, our methodology leverages multi-scale feature fusion within the BEV space, bolstering the precision and robustness of map construction across complex road environments.

In summary, we propose a novel framework for HD map construction with the following key contributions.

\begin{itemize}
   \item   We develop an advanced view transformation module that dynamically links image features to road surface height distributions, significantly enhancing spatial comprehension by integrating seamlessly with attention-based neural networks.
   \item   We introduce a foreground-background separation network optimized for road environments, using self-supervised learning to improve road feature extraction and remove irrelevant background elements. Besides, a coupled multi-scale feature fusion mechanism is proposed to  achieve a more robust BEV representation.
   \item  Extensive experiments on two challenging HD map construction datasets validate the effectiveness of our framework, demonstrating robust performance and suitability across diverse driving scenarios.    
\end{itemize}

\section{Related Work}
\label{sec:relatedwork}

\textbf{Online HD Map Construction.} 
The construction of online HD map from visual perception has recently become a focal point of interest and has seen significant advancements. This process fundamentally involves extracting image features from surround-view inputs. Notably, PETRv2~\cite{liu2023petrv2} adopted a sparse query-based method, initially derived from the 3D object detection field, to create BEV segmentation maps and delineate 3D lanes. This method integrates advanced object recognition frameworks to enable comprehensive environmental modeling. In contrast, BEV-based methods~\cite{li2022hdmapnet, MapTR, maptrv2, liu2022vectormapnet, Yuan_2024_streammapnet} leveraged estimated depths or attention layers to project image features into BEV space, addressing the inherent limitations of pixel coordinate systems in capturing depth information. For instance, methods employing depth series~\cite{huang2022bevpoolv2, maptrv2, philion2020lift} enhanced the transformation from 2D to 3D coordinates by constructing depth profiles directly from image features. Similarly, techniques using attention series~\cite{chen2022efficient, vaswani2017attention} exploited the robust capabilities of attention mechanisms to contextualize 2D image features within a 3D spatial framework.
Furthermore, several methods such as BeMapNet~\cite{BeMapNet}, PivotNet~\cite{ding2023pivotnet}, and MapVR~\cite{MapVR} focused on refining the decoder by integrating prior geometric information about map elements, aiming to achieve more accurate and smoother detection outcomes. Concurrently, some other algorithms~\cite{yu2023scalablemap} typically required additional modules to enhance the output features in the BEV space.

\textbf{Height Modeling for 3D Detection.} 
Height modeling, originally introduced in the 3D object detection domain, has seen limited advancement over the years. Initially, BEVFormer~\cite{li2022bevformer}  implemented the concept of height implicitly by setting reference points at various predefined heights, thus beginning the exploration into vertical dimension modeling. Subsequently, BEVHeight~\cite{yang2023bevheight} presented a method that models object heights by predicting the heights of image pixels and then translating these features into BEV space using geometric transformations. This approach has demonstrated considerable effectiveness in roadside scenarios~\cite{ye2022rope3d, yu2022dair}, where its performance strongly depends on the precision of the camera's installation height. Furthering this progression, HeightFormer~\cite{wu2023heightformer}  expanded the practical application to scenarios adjacent to vehicles and explicitly incorporates the concept of height into its modeling framework. 
Different from these previous works, we make the first attempt that extends the application of height models for HD map construction by developing an explicit height prediction model.

\section{Method}

\begin{figure*}[ht]
  \centering  
  \includegraphics[width=0.78\linewidth]{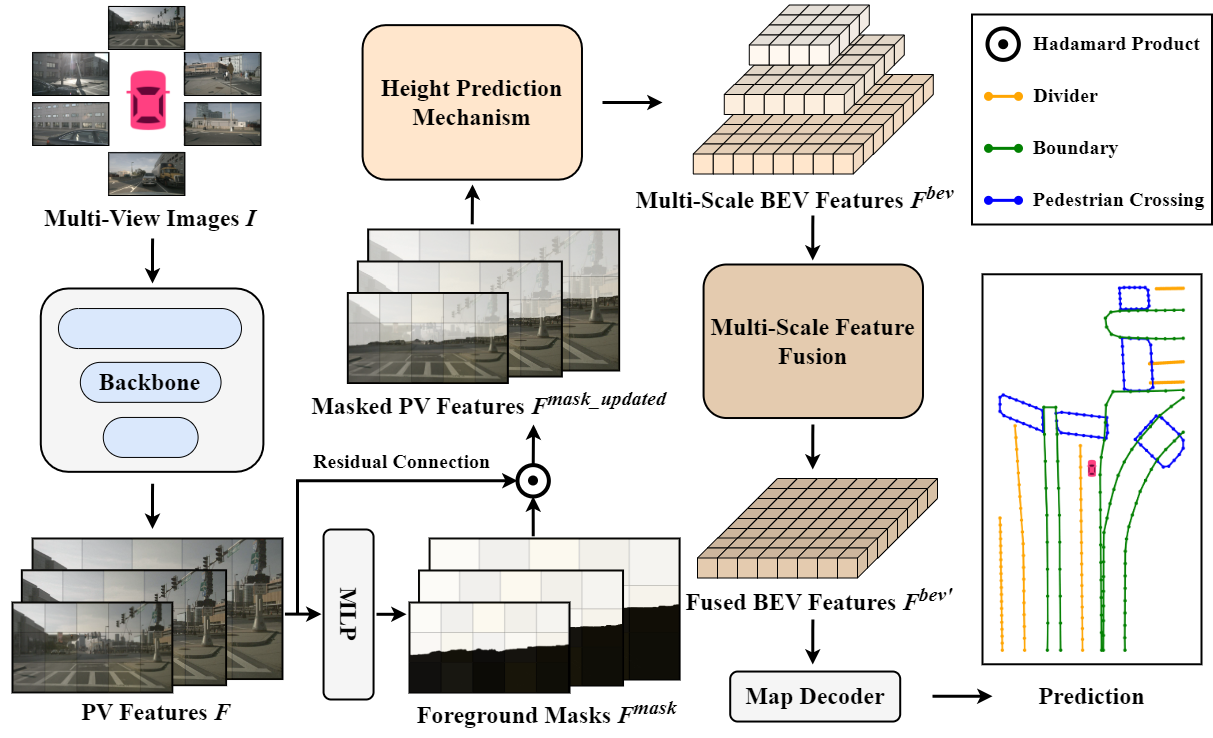}
  \caption{The overview of our proposed HeightMapNet. Our method consists of three main components: a foreground-background separation network that emphasizes roadway features, a height prediction mechanism for PV-to-BEV transformation, and a  multi-scale feature fusion module that enhances BEV representation.}
  \label{fig1}
  \vskip -0.1in
\end{figure*}

\subsection{Overview}
 Figure \ref{fig1} illustrates the workflow  of our approach. It initiates with a feature encoder that extracts multi-scale PV features $F = \{F_1, F_2, \ldots, F_s\}$ (where $s$ is the number of scales) from the raw images $I$. These PV features undergo refinement in a foreground-background separation network, which effectively discriminates between road elements and non-road elements, enhancing the purity of the feature signals. Following this, a height prediction mechanism facilitates the transformation of features from traditional PV to BEV. This transformation significantly improves perceptual accuracy through a  comprehensive spatial representation of the scene. After that,  a multi-scale feature fusion technology is applied, which  enhances the model adaptability to complex scenarios by integrating BEV features captured at different scales. Ultimately, a feature decoder transforms these processed features into a vectorized scene representation. This representation meticulously depicts key static road elements, including divider (div.), boundary (bou.), and pedestrian crossing (ped.), providing a precise reflection of actual road conditions.

\subsection{Foreground-Background Separation}

In the processing of image features, most existing  studies do not sufficiently discriminate between  foreground road elements and non-essential background elements. This inclusion of background information renders the model vulnerable to irrelevant data, potentially impairing its processing efficacy. To address this problem, we  develop a self-supervised foreground-background separation network, leveraging projection relationships to enhance model focus on pertinent road elements. This network is engineered to produce precise foreground masks that effectively attenuate background information not pertinent to the road.

As depicted in Figure \ref{fig1}, the foreground-background separation network receives multi-scale PV features $F_i$  as inputs and processes them through a streamlined multi-layer perceptron (MLP). This MLP generates corresponding foreground masks $F^{mask}_i$, which are then integrated with the original PV features using residual connection~\cite{he2016deep}. This integration employs the Hadamard product to adjust the positional intensity of the feature maps meticulously. The processed data are subsequently added back to the original PV features, enriching them with confidence information. This strategy significantly fortifies the feature set, rendering it more robust against potential disruptions. The operational flow of this process is detailed as follows:
\begin{gather}
F^{mask}_i = \text{Sigmoid}(\text{Conv}(\text{ReLU}(\text{Conv}(F_i)))) \label{eq:mask}, \\
F^{mask\_updated}_i = F_i + F_i \odot F^{mask}_i, ~i=1,\ldots, s. \label{eq:mask_updated}
\end{gather}
Moreover, we introduce a self-supervised methodology for generating ground truths, anchored on geometric projection relationships, to accurately define foreground masks. 
In particular, our procedure initiates with the establishment of uniformly distributed reference points within a designated range (within the range of $[-2.0m, 2.0m]$) in the BEV space. Utilizing the camera's intrinsic and extrinsic parameter matrices, these points are then geometrically projected from the 3D space onto the 2D image plane. Features within the scope of these projections are classified as foreground, encompassing the road and nearby critical elements. In contrast, areas outside these projections are designated as background, typically including non-road elements like the sky.

After processing through the foreground-background separation network, the refined image features achieve a heightened focus on the road and essential adjacent foreground elements. These foreground-delineated features $F^{mask\_updated}_i$ are subsequently employed by the height prediction mechanism, enhancing the effectiveness of the perspective transformation process. 

\subsection{Height Prediction Mechanism}

\begin{figure}[t]
  \centering
  \includegraphics[width=\linewidth]{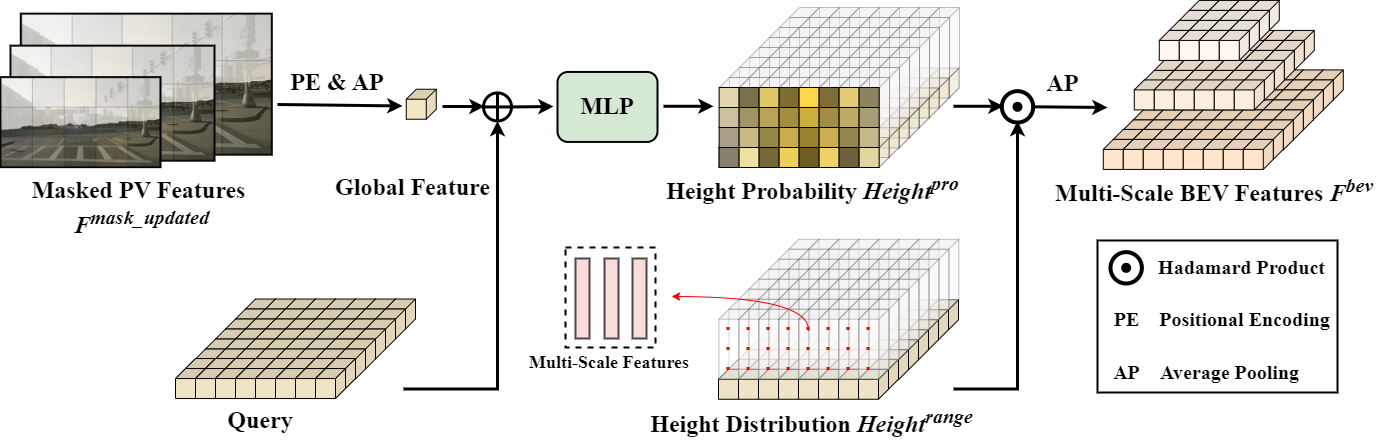}
  \caption{Illustration of height prediction mechanism.}
  \label{fig3}
  \vskip -0.1in
\end{figure}

The perspective transformation in BEV-based frameworks has traditionally been considered an ill-posed problem, often addressed through complex depths or attention mechanisms to directly generate transformed BEV features. Although effective, this approach complicates the model's architecture and obscures its interpretability.  To overcome these limitations, we introduce a novel height prediction mechanism that leverages prior knowledge of height distributions, thereby transforming the task from direct BEV feature generation to estimating height distribution probabilities. This modification substantially alleviates the learning complexity associated with perspective transformations within the model. The architecture of this network is depicted in Figure \ref{fig3}.

In our approach, the masked PV features first undergo preliminary positional encoding $\mathbb{PE}$ and average pooling $\mathbb{AP}$ operations. These steps are designed to abstract the image features globally, establishing a foundational layer for the subsequent modeling of height distribution probabilities. Following this abstraction, the globally-informed image features are integrated into a predefined query, aiming to achieve dynamic initialization. 
The final 
height distribution probabilities, denoted as $Height^{pro}$, is then computed using a straightforward MLP. In the implementation, height distribution probabilities of the \textbf{highest-level} image features are shared to simplify the model.
\begin{equation}
\small
    Height^{pro} = \text{MLP}(\mathbb{AP} (F^{mask\_updated}_i+\mathbb{PE}) + Query), ~i=1.
 \label{eq:height_pro}
\end{equation}
Furthermore, the height prediction mechanism employs a spatial sampling strategy analogous to that used in the foreground-background separation network. This strategy leverages predefined reference points $P^{bev}_{3d}$ proximal to the road, viewed under the BEV space. By projecting these points using the camera’s intrinsic $\mathbb{I}$ and extrinsic $\mathbb{K}$ parameters into the PV space, corresponding features are effectively captured across multiple scales, 
denoted as $Height_i^{range}$.
\begin{equation}
Height_i^{range} = \text{Sampling}_i (\mathbb{I} \cdot \mathbb{K} \cdot P^{bev}_{3d}), ~i=1,\ldots, s. \label{eq:height_range}
\end{equation}
These captured image features, informed by height distribution probabilities, then undergo weighted pooling along the Z-axis. This targeted pooling refines the features proximate to the road, culminating in the derivation of the final BEV features $F_i^{bev}$.
\begin{equation}
\small
F_i^{bev} = \mathbb{AP}(Height^{pro} \odot Height_i^{range}), ~i=1,\ldots, s. \label{eq:F_bev}
\end{equation}
In summary, the architecture and implementation of the height prediction mechanism not only streamline and guide the perspective transformation process but also ensure that the resultant multi-scale BEV features are acutely focused on the road. This focus provides a detailed and comprehensive representation of the surrounding road elements. Through this mechanism, the model successfully transitions from processing PV features to BEV features, setting the stage for the subsequent multi-scale feature fusion module.

\subsection{Multi-Scale Feature Fusion}\label{Multi-Scale Feature Fusion}

\begin{figure}[t]
  \centering
  \includegraphics[width=\linewidth]{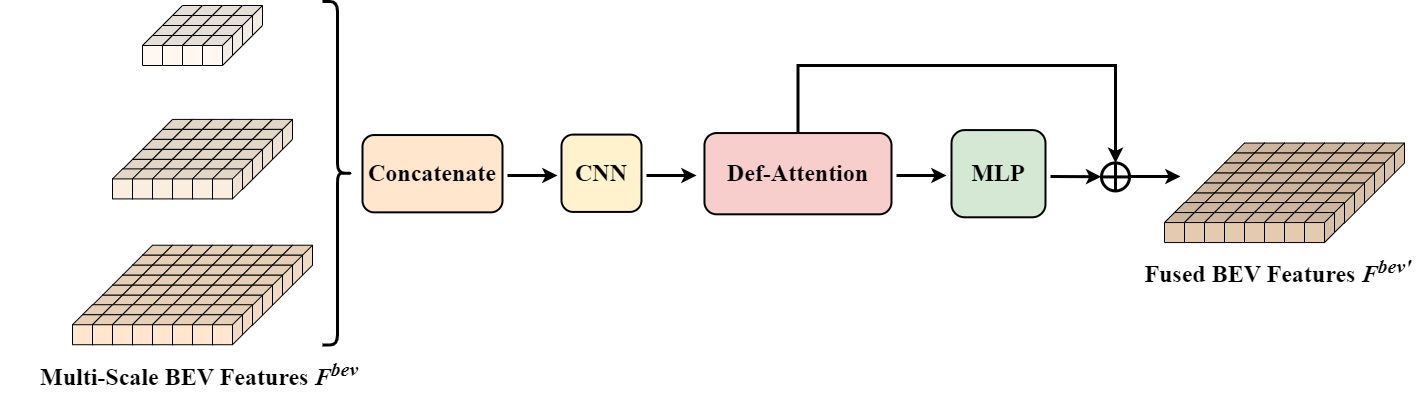}
  \caption{Illustration of multi-scale feature fusion.}
  \label{fig4}
  \vskip -0.1in
\end{figure}

While existing BEV-based methods typically rely on single-layer image features to enhance computational efficiency, this approach is not always the optimal choice for distinguishing different elements in HD maps. Notably, distant elements, which pose greater capture challenges, may benefit from the large-resolution features present in lower layers, whereas larger, nearby elements are more aptly captured by higher layers. To overcome this limitation, inspired by previous studies~\cite{li2022bevformer,liu2022vectormapnet}, we introduce a multi-scale feature fusion module at the BEV stage that orchestrates the integration of features across different scales.

The architecture of  our fusion module  is illustrated in Figure \ref{fig4}.  As shown,  multi-scale BEV features are  coarsely integrated via feature concatenation. This preliminary integration serves as the foundation for further processing by a specially tailored convolutional neural network (CNN). This CNN is designed to enhance the representational capacity of the features and deepen the model’s understanding of spatial dynamics, enabling a more nuanced interpretation of the road environment. 

Furthermore, the integration of a deformable attention mechanism~\cite{xia2022vision} equips the model with an adaptive capability to selectively enhance crucial feature areas pivotal for the final inference. This mechanism dynamically adjusts the model’s focus across various feature regions, ensuring that the feature fusion process is not only more targeted but also efficient. Following the fusion process, the application of an MLP meticulously refines the integration of features. The architecture, utilizing a series of linear layers augmented with strategically placed residual connections, ensures that the resulting BEV features, denoted as $F^{bev'}$, are both enriched and stabilized. This sequence is depicted as follows, where $\mathbb{DA}$ symbolizes the deformable attention mechanism.
\begin{equation}
\begin{split}
F^{bev'} = &\text{MLP}(\mathbb{DA}(\text{Conv}(\text{Concat}(F_1^{bev}, \ldots, F_s^{bev})))) \\
& + \mathbb{DA}(\text{Conv}(\text{Concat}(F_1^{bev}, \ldots, F_s^{bev}))). \label{eq:F_bev_prime}
\end{split}
\end{equation}

\subsection{Training Loss}

HeightMapNet implements an end-to-end training methodology, finely tuned through instance-level and point-level assignments. The training framework is anchored by four pivotal components of loss: classification loss $L_{cls}$, point-to-point loss $L_{pos}$, edge direction loss $L_{dir}$, and mask loss $L_{mask}$, represented as follows:
\begin{gather}
L = \lambda L_{cls} + \alpha L_{pos} + \beta L_{dir} + \gamma L_{mask}. \label{eq:Loss1}
\end{gather}
The losses $L_{cls}$, $L_{pos}$, and $L_{dir}$ are configured in alignment with the established protocols of MapTR ~\cite{MapTR}, ensuring consistency with well-regarded benchmarks. The mask loss $L_{mask}$, specifically designed to enhance the foreground-background separation network’s focus on the road, is quantified using the Manhattan distance between corresponding feature masks. This is mathematically expressed as:
\begin{equation}
L_{mask} = \sum_i D_{\text{Manhattan}}(F^{mask}_i, F^{mask\_GT}_i), \label{eq:Loss2}
\end{equation}
where $F^{mask}_i$, $F^{mask\_GT}_i$ represent the predicted foreground mask and the ground truth foreground mask of the PV features, respectively.

\section{Experiment}
\subsection{Experimental Setup}\label{Sec:setup}

To rigorously evaluate the performance of HeightMapNet in constructing HD maps, comprehensive experiments were conducted on the nuScenes  and  the Argoverse 2. The nuScenes dataset~\cite{Caesar_Bankiti_Lang_Vora_Liong_Xu_Krishnan_Pan_Baldan_Beijbom_2020} comprises 1,000 driving scenes, sourced from urban environments in Boston and Singapore. Each scene includes a 20-second video with key samples annotated at a frequency of 2 Hz, capturing panoramic views using six surround-view cameras mounted on the vehicle. 
The Argoverse 2 dataset~\cite{Argoverse2}  features 20,000 sequences, each lasting 30 seconds, recorded at a high sampling rate of 10 Hz. This dataset employs seven surround-view cameras, offering an exceptional density of data and enhanced temporal continuity.

For the quantitative evaluation, we adopted the chamfer distance $CD$ as a metric to measure the degree of correspondence between the model predictions and the actual road conditions. This metric assesses the average precision by calculating the chamfer distance between two sets of points, $A = \{a_1, a_2, \ldots, a_m\}$ and $B = \{b_1, b_2, \ldots, b_n\}$, where the specific computation method is:
\begin{gather}
\small
CD(A, B) = \frac{1}{|A|} \sum_{a \in A} \min_{b \in B} \| a - b \|^2 + \frac{1}{|B|} \sum_{b \in B} \min_{a \in A} \| b - a \|^2. \label{eq:L}
\end{gather}
The overall average precision $AP$ is subsequently calculated by aggregating values under various chamfer distance thresholds $T = \{0.5, 1.0, 1.5\}m$, represented by the formula: $AP = \frac{1}{|T|} \sum_{\tau \in T} AP_\tau$.

\subsection{Implementation Details}

Our approach extends MapTR~\cite{MapTR}, leveraging its feature encoder and decoder to detect map elements from BEV perspectives. Notably, our method is designed with modularity in mind, facilitating adaptation to alternative BEV-based map detectors.
The perceptual field is defined with an X-axis range of $[-15.0\,m, 15.0\,m]$, and a Y-axis range of $[-30.0\,m, 30.0\,m]$, tailored to capture a comprehensive spatial context.

The model's training was initiated from scratch, utilizing a single NVIDIA A100 GPU. Optimization was conducted using the AdamW~\cite{loshchilov2017decoupled} optimizer, characterized by a weight decay of 1.25e-3, a batch size of 8, and an initial learning rate of 3e-4. The training protocol included 24 complete epochs, aligning with the methodologies employed by previous models~\cite{MapTR, maptrv2}. For adaptations to the Argoverse 2 dataset, modifications included a revised weight decay of 0.01 and a reduced learning rate of 1.5e-4, with the training duration limited to 6 epochs. Except for these specified adjustments, all other training parameters were maintained consistent with those used for MapTR~\cite{MapTR}.

\subsection{Main Results}
\begin{table*}[ht]
  \caption{Comparisons on the nuScenes $\texttt{val}$ set. This table compares the performance of different sensors and architectures including Camera (C), Lidar (L), EfficientNet-B0~\cite{tan2019efficientnet} (EB0), PointPillars~\cite{lang2019pointpillars} (PPs), and ResNet50~\cite{he2016deep} (R50). $\dagger$ means the result is reproduced with public code, and $*$ indicates that HeightMapNet is extended by incorporating the one-to-many loss and BEV loss from MapTRv2.}
  \label{t1}
  \centering
  \small
  \begin{tabular}{lccccccc}
    \toprule
    Methods & Modality & Backbone & Epochs & $AP_{div.}$\text{$\uparrow$} & $AP_{bou.}$\text{$\uparrow$} & $AP_{ped.}$\text{$\uparrow$} & 
    $mAP$\text{$\uparrow$}\\
    \midrule
    HDMapNet~\cite{li2022hdmapnet} & C & EB0 & 30 & 0.217 & 0.330 & 0.144 & 0.230\\
    HDMapNet~\cite{li2022hdmapnet} & C \& L & EB0 \& PPs & 30 & 0.296 & 0.467 & 0.163 & 0.310\\
    VectorMapNet~\cite{liu2022vectormapnet} & C & R50 & 110 & 0.473 & 0.393 & 0.361 & 0.409\\
    VectorMapNet~\cite{liu2022vectormapnet} & C \& L & R50 \& PPs & 110 & 0.505 & 0.475 & 0.376 & 0.452\\
    MapTR$^\dagger$~\cite{MapTR} & C & R50 & 24 & 0.512$^\dagger$ & 0.531$^\dagger$ & 0.493$^\dagger$ & 0.512$^\dagger$\\
    PivotNet~\cite{ding2023pivotnet} & C & R50 & 24 & 0.565 & 0.601 & 0.562 & 0.576\\    
    MapVR~\cite{MapVR} & C & R50 & 110 & 0.618 & 0.594 & 0.550 & 0.588\\    
    MapTRv2$^\dagger$~\cite{maptrv2} & C & R50 & 24 & 0.608$^\dagger$ & 0.616$^\dagger$ & 0.566$^\dagger$ & 0.597$^\dagger$\\    \midrule
    HeightMapNet & C & R50 &24 & \textbf{0.628} & 0.604 & 0.543 & 0.591\\
    HeightMapNet$^*$ & C & R50 & 24 & 0.626 & \textbf{0.622} & \textbf{0.604} & \textbf{0.617}\\
    \bottomrule
  \end{tabular}%
  \vskip -0.1in
\end{table*}

\begin{table}[ht]
  \caption{Comparisons on the nuScenes $\texttt{val}$ set under tighter $\{0.2, 0.5, 1.0\}m$ setting. $\dagger$ means the result is reproduced with public code, and $*$ indicates that HeightMapNet is extended by incorporating the one-to-many loss and BEV loss from MapTRv2. Besides, the results of VectorMapNet, HDMapNet and PivotNet are directly cited from~\cite{ding2023pivotnet}.}
  \label{nuScenes val set tighter}
  \centering
  \resizebox{\linewidth}{!}{%
  \begin{tabular}{lcccccc}
    \toprule
    Methods  & Epochs & $AP_{div.}$\text{$\uparrow$} & $AP_{bou.}$\text{$\uparrow$} & $AP_{ped.}$\text{$\uparrow$} & 
    $mAP$\text{$\uparrow$} & FPS$\uparrow$\\
    \midrule
    VectorMapNet~\cite{liu2022vectormapnet}  &110 &0.272 &0.184 &0.182 &0.213 &--\\
    HDMapNet~\cite{liu2022vectormapnet}   &30 &0.283 &0.326 & 0.071 &0.227 &--\\
    MapTR$^\dagger$~\cite{MapTR}  & 24 & 0.330$^\dagger$ & 0.298$^\dagger$ & 0.256$^\dagger$ & 0.295$^\dagger$ & \textbf{9.1}$^\dagger$\\
    MapTRv2$^\dagger$~\cite{maptrv2}  &24 & 0.428$^\dagger$ & 0.367$^\dagger$ & 0.342$^\dagger$ & 0.379$^\dagger$ & 9.0$^\dagger$\\   
    PivotNet~\cite{ding2023pivotnet}  &24 &0.414 &0.398 &0.343 &0.385 &--\\
    \midrule
    HeightMapNet  & 24 & 0.442 & 0.353 & 0.303 & 0.366 & 7.8\\
    HeightMapNet$^*$  & 24 & \textbf{0.458} & \textbf{0.373} & \textbf{0.364} & \textbf{0.398} & 6.7\\
    \bottomrule
  \end{tabular}
  }
\end{table}

\subsubsection{Results on the nuScenes $\texttt{val}$ Set}
The experimental results of HeightMapNet on the nuScenes $\texttt{val}$ set are systematically presented in Table \ref{t1}. This table not only documents the performance metrics of HeightMapNet but also juxtaposes these results with those of other leading algorithms tested under equivalent conditions, providing a comprehensive comparative analysis. 
Specifically, HeightMapNet  registers  an overall increase of 7.9\% in mean Average Precision ($mAP$) than the baseline MapTR~\cite{MapTR},  affirming its enhanced accuracy in map element detection. 
A detailed examination of the results reveals that HeightMapNet is particularly adept at detecting divider, achieving an 11.6\% boost. Furthermore, by incorporating  the one-to-many loss and BEV loss from MapTRv2~\cite{maptrv2}, our method outpeforms all the compared methods by a considerable margin.

Following~\cite{ding2023pivotnet}, we also give detection accuracy under three tighter thresholds of $\{0.2, 0.5, 1.0\}$m in Table~\ref{nuScenes val set tighter}, where  a prediction is treated as true positive (TP)
only if the distance between prediction and ground-truth is less than the specified threshold.
As shown, compared to the baseline MapTR, our HeightMapNet achieves 7.1\% higher mAP. Compared to the existing state-of-the-art PivotNet~\cite{ding2023pivotnet},  HeightMapNet$^*$  enjoys 1.3\% higher mAP. These again verify the effectiveness of proposed solution.

Table \ref{nuScenes val set tighter} also provides the inference speed of our approach and baselines. All numbers in the column of FPS are evaluated on a same device for a fair comparison. As shown, our method improves detection accuracy without too much efficiency sacrifice. Specifically, HeightMapNet achieves 7.8 FPS, while the corresponding value for MapTR~\cite{MapTR} is 9.1. 

\subsubsection{Results on the nuScenes $\texttt{Test}$ Set}
To further substantiate the efficacy of HeightMapNet, we extended our evaluation to include the $\texttt{test}$ set, achieving a broader validation as detailed in Table \ref{nuScenes test set}. The results  affirm HeightMapNet's substantial advancements over MapTR across various chamfer distance thresholds. 
As can be seen, under the general threshold of $\{0.5, 1.0, 1.5\}m$, both MapTR and HeightMapNet display significant improvements in $mAP$ on the nuScenes $\texttt{test}$ set, with each surpassing their respective performances on the $\texttt{val}$ set by over 10\%. This enhancement corroborates the models' effectiveness and reliability in real-world applications. Specifically, MapTR~\cite{MapTR} achieves an $mAP$ of 63.9\% on the $\texttt{test}$ set, whereas HeightMapNet records a superior $mAP$ of 71.0\%.  Besides, our method achieves 6.5\% higher mAP with the tighter $\{0.2, 0.5, 1.0\}m$ setting compared to MapTR~\cite{MapTR}, further verifying the effectiveness of the proposed  components.

\begin{table*}[ht]
  \caption{Comparisons on the nuScenes $\texttt{test}$ set. The input modality utilized is Camera. $\dagger$ means the result is reproduced with public code, and $*$ indicates that HeightMapNet is extended by incorporating the one-to-many loss and BEV loss from MapTRv2. All the results are obtained on a same device, with a training duration of 24 epochs and ResNet50 serving as the backbone.}
  \label{nuScenes test set}
  \centering
  \small
  \begin{tabular}{l|cccc|cccc}
    \toprule
    \multirow{2}{*}[-0.3em]{Methods} & \multicolumn{4}{c|}{\textit{General CD}: $\{0.5, 1.0, 1.5\}m$$\uparrow$} & \multicolumn{4}{c}{\textit{Tighter CD}: $\{0.2, 0.5, 1.0\}m$$\uparrow$}\\
    \cmidrule(lr){2-5} \cmidrule(lr){6-9}
    & $AP_{div.}$ & $AP_{bou.}$ & $AP_{ped.}$ & $mAP$ & $AP_{div.}$ & $AP_{bou.}$ & $AP_{ped.}$ & $mAP$ \\
    \midrule
    MapTR$^\dagger$~\cite{MapTR} & 0.682$^\dagger$ & 0.619$^\dagger$ & 0.617$^\dagger$ & 0.639$^\dagger$ & 0.449$^\dagger$ & 0.359$^\dagger$ & 0.339$^\dagger$ & 0.382$^\dagger$\\
    MapTRv2$^\dagger$~\cite{maptrv2} & 0.762$^\dagger$ & 0.700$^\dagger$ & 0.690$^\dagger$ & 0.717$^\dagger$ & 0.544$^\dagger$ & 0.425$^\dagger$ & 0.401$^\dagger$ & 0.457$^\dagger$\\    
    \midrule
    HeightMapNet & \textbf{0.773} & 0.693 & 0.665 & 0.710 & 0.552 & 0.421 & 0.369 & 0.447\\    
    HeightMapNet$^*$ & 0.772 & \textbf{0.706} & \textbf{0.721} & \textbf{0.733} & \textbf{0.565} & \textbf{0.432} & \textbf{0.422} & \textbf{0.473}\\
    \bottomrule
  \end{tabular}%
\end{table*}
\begin{table}[t]
  \caption{Comparisons on the Argoverse 2 $\texttt{val}$ set. The input modality utilized is Camera. $\dagger$ means the result is reproduced with public code, and $*$ indicates that HeightMapNet is extended by incorporating the one-to-many loss and BEV loss from MapTRv2. "-" indicates that the corresponding results are not available.} 
  \label{t3}
  \centering
  \resizebox{\linewidth}{!}{%
  \begin{tabular}{lccccc}
    \toprule
    Methods & Epochs & $AP_{div.}$\text{$\uparrow$} & $AP_{bou.}$\text{$\uparrow$} & $AP_{ped.}$\text{$\uparrow$} & 
    $mAP$\text{$\uparrow$}\\
    \midrule
    HDMapNet~\cite{li2022hdmapnet}  & 6 & 0.057 & 0.376 & 0.131 & 0.188\\
    VectorMapNet~\cite{liu2022vectormapnet}  & 24 & 0.361 & 0.392 & 0.383 & 0.379\\
    MapTR$^\dagger$~\cite{MapTR}  & 6 & 0.533$^\dagger$ & 0.562$^\dagger$ & 0.509$^\dagger$ & 0.535$^\dagger$\\
    MapVR~\cite{MapVR}  & - & 0.600 & 0.580 & 0.546 & 0.575\\
    MapTRv2$^\dagger$~\cite{maptrv2}  & 6 & 0.704$^\dagger$ & 0.664$^\dagger$ & 0.620$^\dagger$ & 0.663$^\dagger$\\
    \midrule
    HeightMapNet  & 6 & 0.639 & 0.627 & 0.667 & 0.644\\    
    HeightMapNet$^*$  & 6 & \textbf{0.706} & \textbf{0.679} & \textbf{0.627} & \textbf{0.671}\\
    \bottomrule
  \end{tabular}
  }
\end{table}

\subsubsection{Results on the Argoverse 2 $\texttt{Val}$ Set}
Dedicated to comprehensively evaluating the generalization capabilities and performance of the HeightMapNet, we executed extensive comparative experiments on the Argoverse 2 dataset. The outcomes of these experiments are methodically detailed in Table~\ref{t3}. As shown, HeightMapNet achieves an $mAP$ of 64.4\%, marking an overall enhancement of 10.9\% than MapTR~\cite{MapTR}. This considerable improvement not only highlights HeightMapNet's superior performance in processing large-scale datasets but also robustly supports the model’s substantial potential for broader applications in HD map construction tasks. Moreover, our method outperforms other competitors also by a large margin. For instance, the gain over the recent approach MapVR~\cite{MapVR} is up to 6.9\%.

\begin{table}[ht]
  \caption{Ablation study on the nuScenes $\texttt{val}$ set. This ablation study details the incremental addition of various modules to the baseline. Auxiliary Loss in configuration \#5 indicates that the one-to-many loss and BEV loss from MapTRv2 have been added to the basic HeightMapNet.}
  \label{Ablation study on the nuScenes val set}
  \centering
  \resizebox{\linewidth}{!}{%
  \begin{tabular}{clcccc}
    \toprule
    \# & Methods & $AP_{div.}$\text{$\uparrow$} & $AP_{bou.}$\text{$\uparrow$} & $AP_{ped.}$\text{$\uparrow$} & 
    $mAP$\text{$\uparrow$}\\
    \midrule
    1 & Baseline (MapTR) & 0.512 & 0.531 & 0.493 & 0.512\\
    2 & + Height Prediction Mechanism & 0.571 & 0.566 & 0.516 & 0.551\\
    3 & + Foreground-Background Separation & 0.570 & 0.567 & 0.555 & 0.564\\
    4 & + Multi-Scale Feature Fusion (HeightMapNet) & \textbf{0.628} & 0.604 & 0.543 & 0.591\\
    5 & + Auxiliary Loss (HeightMapNet$^*$) & 0.626 & \textbf{0.622} & \textbf{0.604} & \textbf{0.617}\\
    \bottomrule
  \end{tabular}
  }
  \vskip -0.1in
\end{table}

\subsection{Ablation \& Analysis}
\subsubsection{Impact of Each Component}
Here, we present an analysis of the critical components of our model, elucidating their individual contributions to performance enhancement. As outlined in Table \ref{Ablation study on the nuScenes val set}, we begin with MapTR as the baseline configuration \#1, with subsequent integrations of each module to assess their impacts.

\textbf{Height Prediction Mechanism.} The integration of the height prediction mechanism into MapTR led to noticeable improvements in detection accuracy for all the three elements. The most substantial gain was observed in divider detection. This enhancement is attributed to the module’s precise capture of height information of road elements relative to the ego vehicle's coordinate system. Such precision significantly refines the representativeness of the BEV features transformed from this data, thereby enhancing the model’s inferential accuracy.

\textbf{Foreground-Background Separation.} Adding the foreground-background separation focused the model’s attention more sharply on elements near the road surface before perspective transformation. This change is reflected in the improved detection precision for pedestrian crossing noted in ablation experiments. Conversely, no significant accuracy changes in accuracy for divider and boundary detection were noted, likely due to two reasons: 1) Background elements sometimes provide indirect contextual information critical for identifying dividers. Masking these elements may deprive the model of essential environmental cues. 2) In certain conditions, the texture and color contrasts in the background help in distinguishing dividers and boundaries. Removing these elements might diminish the visual distinctiveness of these two elements, impacting detection accuracy adversely.

\textbf{Multi-Scale Feature Fusion.} The adoption of this module yielded a 5.8\% increase in divider detection accuracy and a 3.7\% improvement for boundary. This significant uplift is credited to the module’s capacity to synthesize both coarse and fine-grained features, thus maintaining a global perspective while enhancing local detail discernment—essential for accurately identifying the intricate and elongated shapes of divider and boundary. However, a slight reduction of 1.2\% in pedestrian crossing detection accuracy was observed, which could stem from the concentrated feature distribution typical of pedestrian areas. In such instances, the fusion of multi-scale features might introduce unnecessary noise, disrupting the model’s capacity to distinguish and learn crucial characteristics.

\begin{table}[t]
  \caption{Ablation study on different height values. This ablation study delineates the impact of varying the number of road surface height sampling points on the performance of HeightMapNet. 
  }
  \label{ablation of height value}
  \centering
  \small
  \begin{tabular}{ccccc}
    \toprule
    Height Values & $AP_{div.}$\text{$\uparrow$} & $AP_{bou.}$\text{$\uparrow$} & $AP_{ped.}$\text{$\uparrow$} & 
    $mAP$\text{$\uparrow$}\\
    \midrule
    8 & 0.619 & 0.593 & \textbf{0.552} & 0.588\\
    12 & \textbf{0.628} & \textbf{0.604} & 0.543 & \textbf{0.591}\\
    16 & 0.618 & 0.600 & 0.549 & 0.589\\
    \bottomrule
  \end{tabular}
  \vskip -0.1in
\end{table}
\subsubsection{Parameter Investigation}
In this section,  the effect of introduced parameters to the proposed approach is analysed. All the results are obtained on the nuScenes $\texttt{val}$ set, with a training duration of 24 epochs and ResNet50 serving as the backbone.
\begin{figure*}[t]
  \centering
  \includegraphics[width=0.83\textwidth]{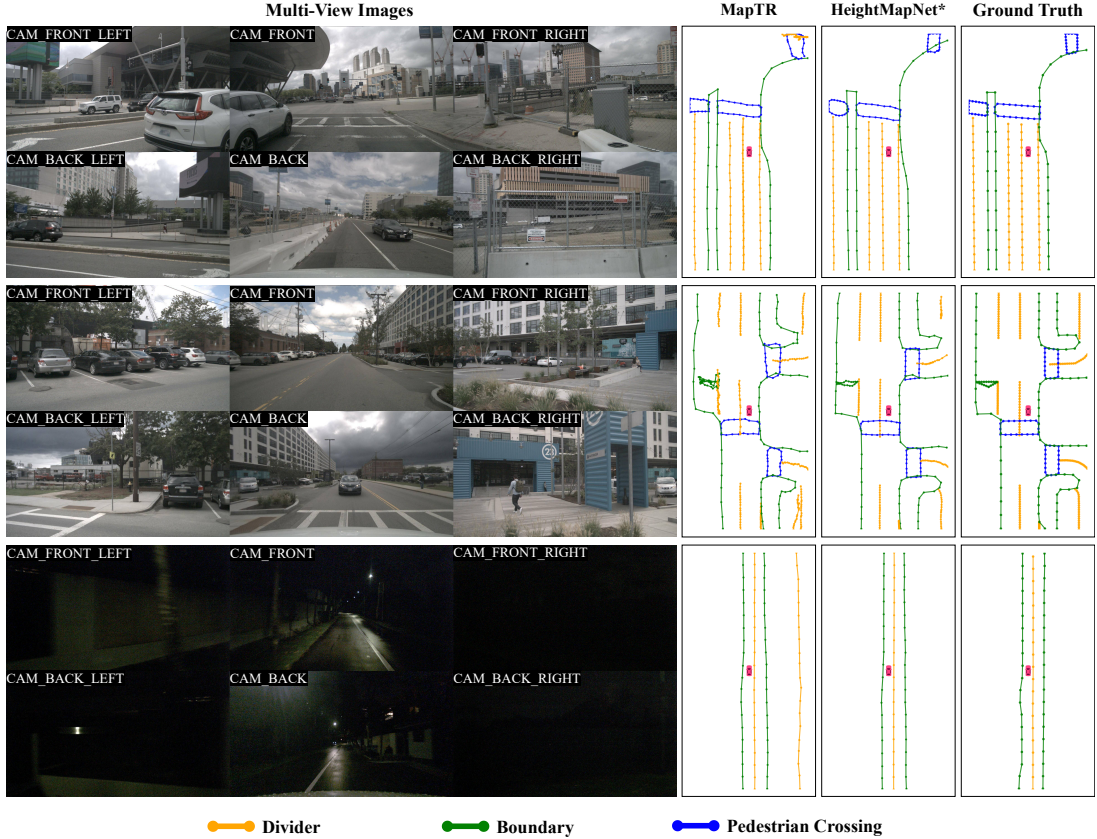}
      \caption{Visualizations on the nuScenes $\texttt{val}$ set under different weather conditions.}
  \label{fig6}
  \vskip -0.15in
\end{figure*}

\textbf{The Number of Height Values.}
To assess the influence of different numbers of height sampling points near the road surface within the height prediction mechanism, various height values were systematically tested. According to the results presented in Table \ref{ablation of height value}, HeightMapNet demonstrates superior performance when the number of height sampling points is set to 12. This outcome highlights the critical role of optimal sampling point configuration in maximizing the model's efficacy.

\textbf{The Values of Hyper-parameter $\gamma$.}
To validate the impact of self-supervised loss weights on the performance of a foreground-background separation network, we experimentally adjusted the  $\gamma$ parameter. As illustrated in Table \ref{ablation of gamma}, HeightMapNet achieves optimal performance when the self-supervised loss weight is set to 1.0. This finding underscores the significance of appropriate loss weighting in enhancing model efficacy.

\begin{table}[t]
  \caption{Ablation study on different $\gamma$. This ablation study investigates the effects of varying self-supervised loss weights on the performance of HeightMapNet.  }
  \label{ablation of gamma}
  \centering
  \small 
  \begin{tabular}{ccccc}
    \toprule
    $\gamma$ & $AP_{div.}$\text{$\uparrow$} & $AP_{bou.}$\text{$\uparrow$} & $AP_{ped.}$\text{$\uparrow$} & 
    $mAP$\text{$\uparrow$}\\
    \midrule
    0.01 & 0.606 & 0.592 & 0.546 & 0.581\\
    0.1 & 0.615 & 0.591 & 0.552 & 0.586\\
    1.0 & \textbf{0.628} & \textbf{0.604} & 0.543 & \textbf{0.591}\\
    10.0 & 0.613 & 0.604 & \textbf{0.554} & 0.590\\
    \bottomrule
  \end{tabular}
  \vskip -0.1in
\end{table}

\subsection{Visualization Results}
To elucidate the practical impact of our method in real-world HD map construction tasks, we  visualize the model’s prediction outcomes in Figure \ref{fig6}. As shown, our method maintains better performance than baselines in different environments. Even at night,
the map near the vehicle closely matches the ground truth.  More qualitative results on the nuScenes $\texttt{val}$ set, including both successful and fail cases, can be found in the supplementary material.

\section{Conclusion}
In this paper, we have developed a novel perspective transformation framework, HeightMapNet, which innovatively incorporates height priors. This model employs a foreground-background separation to meticulously focus on road surface features. Furthermore, by integrating multi-scale features within the BEV space, our approach effectively maximizes the utilization of spatial geometric information. Experimental validations underscore the robust performance of HeightMapNet, showcasing its substantial potential for practical applications.

\textbf{Limitations and Future Work.} While the self-supervised learning strategy utilized in the foreground-background separation network minimizes reliance on dataset annotations, the integration of direct label guidance through supervised learning could provide clearer learning targets. This enhancement is anticipated to further improve both the accuracy and reliability of the model. Looking ahead, future investigations should consider incorporating temporal information to extend the model’s performance, particularly in dynamic and complex environments.

{\small
\bibliographystyle{ieee_fullname}
\bibliography{main}
}

\end{document}